\documentclass[10pt,twocolumn,letterpaper]{article}

\usepackage{times}
\usepackage{epsfig}
\usepackage{graphicx}
\usepackage{amsmath}
\usepackage{amssymb}
\usepackage{url}
\usepackage{amssymb}
\usepackage{algorithmic}
\usepackage{amstext}
\usepackage{amsmath}
\usepackage{amsthm}
\usepackage{multicol}
\usepackage{pslatex}
\usepackage{fancyhdr}

\usepackage{authblk}

\newcommand{\functionName}{Pish}
\date{Submitted Preprint}

\begin{document}

\title{A Note on Deepfake Detection with Low-Resources}

\author[1]{Piotr Kawa}
\author[1]{Piotr Syga}

\affil[1]{
Wroclaw University of Science and Technology\\
Wybrzeze Stanislawa Wyspianskiego 27, 50-370 Wroclaw, Poland
}

\affil[ ]{\textit {piotr.kawa@student.pwr.edu.pl, piotr.syga@pwr.edu.pl}}

\maketitle
\thispagestyle{empty}

\begin{abstract}
Deepfakes are videos that include changes, quite often substituting face of a portrayed individual with a different face using neural networks. Even though the technology gained its popularity as a carrier of jokes and parodies it raises a serious threat to ones security -- via biometric impersonation or besmearing.  In this paper we present two methods that allow detecting Deepfakes for a user without significant computational power. In particular, we enhance MesoNet~\cite{mesonet} by replacing the original activation functions allowing a nearly $1\%$ improvement as well as increasing the consistency of the results. Moreover, we introduced and verified a new activation function --- \functionName{} that at the cost of slight time overhead allows even higher consistency.

Additionally, we present a preliminary results of Deepfake detection method based on Local Feature Descriptors (LFD), that allows setting up the system even faster and without resorting to GPU computation. Our method achieved Equal Error Rate of $0.28$, with both accuracy and recall exceeding $0.7$.
\end{abstract}

\section{Introduction}

The progress in the field of machine learning, in particular deep neural networks, provided tools that automatize tasks that previously required high skills and were highly time--consuming.
One of such tasks is manipulation of facial image, which was earlier possible to perform only by a skilled person, using specialized software. 
Nowadays, thanks to the Deepfake technology it is possible to create face manipulation videos, on a level of detail unachieveable before,  without any sophisticated requirements.

Creation of Deepfake forgeries requires facial images of two persons --- the individual present on the source material and the victim --- person whose face will be added to the video.
Using these two sets, two autoencoders are trained in order to reconstruct the appropriate face --- in result getting expression and illumination of the victim, while keeping facial features of the first person.
Recently, more sophisticated versions of Deepfakes were introduced, with utilization of Generative Adversarial (GAN) networks~\cite{gan}, enhancing the level of detail possible to achieve.
Easy and open access to such technologies leads to many threats that concern various spheres of people's lives.
Deepfakes can be used to synthesize besmearing videos, very often of a pornographic nature. Artificially created fake--news videos, so dangerous in these times, can depict a public figure saying or doing things that might harm their reputation or cause crisis. In addition, the possibility of forging a video in a transparent way undermines the validity of the strength of recordings used as the evidence in courts of law. Naturally, during trials one can be cleared of charged, after detailed examination of the video, yet the process is time--consuming and in the case of public figures, the trust of general opinion may be irreparably tarnished. In such cases a quick and easy method, that allows one to determine whether the movie is legitimate, cannot be overstated. Providing methods that allow Deepfake detection without significant computational power and without delay is even more important, due to increase in remote communication, including financial and clerical matters. Additionally, Deepfakes can influence the performance of biometric identification methods, providing false positives. The problem is relevant especially in the face of regulations like PSD2~\cite{psd2} that may lead to using video conferences as a method of bank client identification.

State--of--the--art Deepfake detection models based on deep neural networks provide results that may be used in the court of law to examine video evidence. However, their high computational cost makes them unavailable for an individual user. Typically, such scenarios are handled by the use of external services which examine the uploaded data, yet due to highly controversial nature of these videos and costs such external services may not be suitable for everyone wanting to verify the news or clear their name.
Notably, there are other architectures like MesoNet \cite{mesonet}, which despite not providing cutting edge results, are good enough to allow an average citizen to have an access to a tool that allows to determine the legitimacy of the material. \\

Due to the fact that modern Deepfake solutions utilize deep neural networks, one of their main components that highly influences final performance, is an activation function.

By following linear transformation in the neuron, activation function allows neural networks to map non--linear nature of the data.
One of most widely used activation functions are ReLU and LeakyReLU~\cite{relu,leaky_relu} --- both used in MesoNet. Their main advantages include simplicity, yet the effectiveness might be improved.

In the paper we focus on detecting Deepfake videos in a fast and reliable manner, so that users may, on their own, determine whether the presented recording is legitimate, or provide a proof that they became victim of besmearing video, before the damage to the reputation spreads. We investigate the influence of the activation functions used in known architectures as well as using a different approach.

\paragraph{Our contribution} In this paper we present a comparison of the performances achieved by MesoNet Deepfake detection model when trained using different activation functions. Moreover, we present and describe a new activation function --- \functionName{}{}. Furthermore, we show, that the initial performance can be increased by using other functions. In addition, we compare the performance of \functionName{} with state--of--the--art solutions using well--known neural networks like SqueezeNet~\cite{squeeze-net}, DenseNet~\cite{dense-net} and EfficientNet \cite{efficientnet}.
Finally, we show promising results of preliminary experiment concerning use of two local feature descriptors (ORB~\cite{orb} and BRISK~\cite{brisk}) that were previously not taken into account in the task of Deepfake detection.

\section{Previous work}\label{sect:previous-work}

The proposed approaches to Deepfake detection come from various fields, many of them based mainly on computer vision methods like frequency analysis \cite{frequency_analysis} where authors look for marginal inconsistencies, local feature descriptors \cite{local_feature_descriptors} or head pose estimation \cite{head_pose_estimation} exploiting image inconsistencies, when frontal face is embedded into tilted head. The others utilized deep neural networks --- most often performing transfer--learning like XceptionNet \cite{xception-net,faceforensics++}. Using this solution allowed to achieve a total accuracy on low quality samples of 0.7010, however utilizing significant computational power.
A different approach, has been presented in~\cite{mesonet}, where the authors provided a relatively shallow architecture, that allows training and verification of the suspected video, with significantly lower resources required, yet at the cost of accuracy resulting in total accuracy at the level of 0.66. More details on MesoNet is presented in~Sect.~\ref{ssect:meso}.
A novel approach, avoiding the usage of deep learning methods, has been presented in~\cite{local_feature_descriptors}. The authors utilize Local Feature Descriptors(LFD) in order to localize inconsistencies in the video. The paper showed only preliminary results, hence only a part of the popular descriptors was evaluated and the results were not paramount with Equal Error Rate going as high as 0.7 and 0.98 False Rejection Rate with a False Acceptance rate of 0.1. However, tested on various kinds of manipulations, the results of LFD showed greater performance than methods like~\cite{korshunov} in the case of low quality Deepfakes.
For broader analysis of Deepfake detection methods, their vulnerabilities and the datasets used one can refer to~\cite{deepfake_survey_1,deepfake_survey_2}.

Recent years were also the time of the improvement in the field of activation functions. One of the recently introduced solutions, Mish \cite{mish}, showed an increase in accuracies when compared to most of the baseline solutions.

\section{Activation functions} \label{sect:activation-functions}

Many different activation functions were introduced throughout recent years.
Some of them showed a progress in relation to their predecessor. However, they have not replaced baseline solutions as their superiority was not universal enough.
This has partially changed with the introduction of two activation functions --- Swish~\cite{swish} and Mish~\cite{mish}. 
They were evaluated on many various problems and most of the time outperformed baseline solutions.

\begin{figure}[h]
    \begin{center}
      \includegraphics[width=0.9\linewidth]{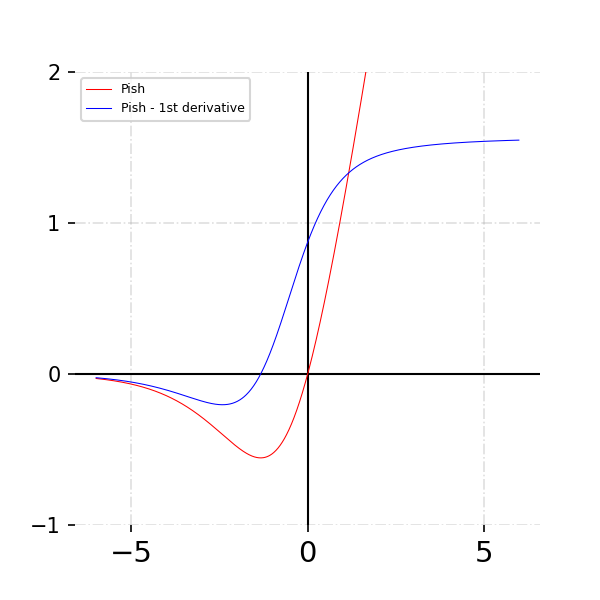}
    \end{center}
      \caption{\functionName{} activation function and its derivative}
    \label{fig:our-activation-function} 
\end{figure}

\functionName{} (Fig.~\ref{fig:our-activation-function}) is a new function proposed by us, whose properties were inspired by the state of the art functions like Swish and Mish. It is defined by $f(x) = x \cdot \text{arctan}(\text{softplus}(x) + \text{sigmoid}(x))$, where $\text{softplus}(x) = \ln(1+e^x)$ and $\text{sigmoid}(x) = \frac{1}{1+e^{-x}}$.
It is a smooth and non--monotonic function, it is characterized by a shift in values near 0. Values of \functionName{} are bounded from below and unbounded at the top, with a range of values of $[\approx -0.56, + \infty)$. Note that the function is continuous, with a smooth derivative 
$f'(x)=\frac{x \big(\frac{e^{-x}}{(e^{-x} + 1)^2} + \frac{e^x}{(1 + e^x)}\big)}{\big(\frac{1}{e^{-x} + 1} + \log(1 + e^x)\big)^2 + 1} + \tan^{-1}\big(\frac{1}{e^{-x} + 1} + \log(1 + e^x)\big)$.
Its performance, together with other aforementioned functions, has been tested on datasets like MesoNet Deepfake detection dataset, Fashion MNIST and CIFAR--10.
Note that it is slightly more computationally demanding than the aforementioned functions (cf.~Sect.~\ref{sect:mesonet-results}), yet shows signs of higher consistency.

\section{MesoNet}\label{ssect:meso}

MesoNet is an architecture of convolutional neural network proposed in~\cite{mesonet} that comes in two variants --- Meso---4 and MesoInception--4.
Based on an input image, the network determines if presented facial image is either Deepfake manipulated or pristine.

Meso--4 begins with 4 blocks, each consisting of a convolutional layer with ReLU activation function, batch normalization and ending with max--pooling layer.
They are followed by a fully--connected layer of 16 neurons with a dropout and ultimate neuron determining validity of an image.
MesonInception--4 comes with inception modules~\cite{inception_layer} in place of two first blocks.

The solution is shallow in comparison to other architectures like ResNet--50 \cite{resnet-50} or XceptionNet \cite{xception-net} that are commonly used in Deepfake detection task. MesoNet solution comes with the number of 27.977 and 28.615 trainable parameters for respectively Meso--4 and MesoInception--4 variants whereas ResNet--50 is composed of over 23.5 million parameters. 
Such difference allows using MesoNet in an environment that does not provide resources needed for the other architectures and determine the legitimacy of the video in shorter time.

The dataset used by Afchar et al. was self--collected. 
It was composed of batches of frames containing different individuals.
The dataset consisted of 12353 training (5103 fake, 7250 pristine) and 7104 test (2845 fake, 4259 pristine) samples.

\section{Deepfake detection experiments}\label{sect:methodology}
MesoInception--4 utilizes two activation functions --- ReLU, in inception and convolution modules and Leaky ReLU in fully connected layer. 
The comparison included ReLU, Leaky ReLU (following the original, the coefficient used was equal to $0.1$), Mish, Swish and \functionName{}. 
In order to determine the influence of various activation functions, in particular \functionName{}, on the performance of Deepfake detection with MesoInception--4 network we performed exhaustive tests. Activations were used in two slots --- the first function was in both inception and convolution modules and the other one was after a fully--connected layer.  
The authors of~\cite{mesonet} have provided the code base that contained networks' implementation, as well as their pretrained weights. For a fair comparison we used, the same parameters whenever they were specified. 
The input data was of size $256\times256\times3$; the optimization was performed using Adam optimizer~\cite{adam}. In addition, training process used a batch size of 75 samples and an adjustable learning rate that decreased every 1000 steps by the factor of ten from $10^{-3}$ down to $10^{-6}$.

Training data underwent data augmentation process including zoom, flips, rotations and color appearance parameter adjustments. Data augmentation details, as well as the number of epochs, were selected by us in the course of the experiment. 

The comparison used the dataset provided by Afchar et al. As no validation set was explicitly separated, it has been extracted by use of 10\% of training images.

This process was done with the respect to the scenes --- in order to prevent data leakage whole scenes were moved. \\

We divided the experiments into 3 stages. First of them concerned training each combination of activation functions pairs (36 different models) for 50 epochs, 
This value was selected in order to achieve comparable computation time to estimated by Afchar et.al.
Using 50 epochs corresponded to 2--3 hours of training time on consumer grade GPU.
Each epoch used all training samples and was ended with a validation of the current state of the network using the whole validation set.
To make sure that the results are reproducible, the stage was repeated on various data splits. \\ 

The next stage of the experiment was done in order to give more details about the efficacy of functions when trained using different learning rates, a common test which was incorporated in works like~\cite{swish, mish}.
We investigated best 5 combinations from the previous stage together with the one originally used in the paper.
Functions were trained using decreasing learning rates starting with $10^{-2}$, $10^{-3}$ and $10^{-4}$. This resulted in the final rates of correspondingly $10^{-5}$, $10^{-6}$ and $10^{-7}$. Each of the 18 models was trained once again on three different splits to provide more reliable results.

The number of training epochs was another detail that differed from the previous stage. Instead of running a training for a fixed number of epochs, process was terminated if the validation accuracy has not improved for three epochs. This aimed to ensure that a model will generalize well and avoid overfitting. \\

The final stage of the experiment was performed on 5 models that provided the best results in the 2nd stage.
Training concerned whole training set, the hyperparameters were selected on the basis of previous stages. 
Models were later tested on testing set. To ensure reliability, process was repeated 5 times using a random data split with a condition of moving whole scene to a single set.

\section{Results}\label{sect:mesonet-results}

Results from the first stage clearly showed the increase in the case of newer activation functions. 13 out of 36 functions' combinations (denoted as convolution activation + fully connected activation e.g. ReLU + Mish) provided better performance than the baseline pair ReLU + Leaky ReLU.
This fact does not particularly implies that these standard functions should be omitted --- the maximum validation accuracy was achieved by ReLU + Swish pair (0.9253). 
\functionName{} was also present among top results --- both in the place of convolution, as well as, the fully connected layers.
The increase in the performance due to usage of Mish function was clearly visible. It was present in the half of the best 10 results, and four out of five best scores.

The best 5 of the combinations with respect to validation accuracy were selected for the next stage --- namely ReLU + Swish, \functionName{} + Mish, Mish + Mish, Swish + Mish, Mish + Swish and ReLU + Leaky ReLU.

\begin{table}[ht!]
\begin{center}
\begin{tabular}{|l|c|c|c|}
\hline
Combination            & LR        & Val. Acc.          & Epochs \\
\hline\hline
        Mish + Mish            & $10^{-2}$      & \textbf{0.8800}   & 16     \\
        \functionName{} + Mish & $10^{-3}$     & 0.8547             & 10     \\ 
        ReLU + Leaky ReLU      & $10^{-3}$     & 0.8447             & 11     \\ 
        ReLU + Swish           & $10^{-3}$     & 0.8375             & 9      \\ 
        Swish + Mish           & $10^{-3}$     & 0.8372             & 10     \\ 
        Mish + Swish           & $10^{-3}$     & 0.8331             & 10     \\
        Mish + Mish            & $10^{-3}$     & 0.8292             & 10     \\
        Swish + Mish           & $10^{-2}$     & 0.8289             & 7      \\ 
        Mish + Swish           & $10^{-4}$    & 0.8289              & 26     \\ 
        ReLU + Swish           & $10^{-2}$    & 0.8097              & 11     \\
        \hline
\end{tabular}
\end{center}
\caption{\label{tab:meso-net-second-stage-results} Maximum validation accuracies scored by the models trained using given learning rates (LR) in the second stage. The epoch number is an average of training durations of all 3 runs.}
\end{table}

Table~\ref{tab:meso-net-second-stage-results} shows 10 best models from the second stage along with their maximum validation accuracies and average number of epochs.
Five of them were selected for the final stage.

\begin{table}[ht!]
    \begin{center} \resizebox{\columnwidth}{!}{
        \begin{tabular}{|l|c|c|}
            \hline
            Model (Learning Rate)          & Accuracy  & Test Accuracy \\
            \hline\hline
            Swish + Mish (0.001)           & 0.9106    & \textbf{0.8718} \\
            \functionName{} + Mish (0.001) & 0.9013    & 0.8689 \\
            ReLU + Leaky ReLU (0.001)      & 0.9052    & 0.8676 \\
            Mish + Mish (0.01)             & 0.9193    & 0.8601 \\
            ReLU + Swish (0.001)           & 0.9252    & 0.8403 \\ 
            \hline
        \end{tabular} }
    \end{center}
    \caption{\label{tab:meso-net-third-stage-results} Accuracies of the best models from the third stage.}
\end{table}

Table~\ref{tab:meso-net-third-stage-results} contains results of mentioned models. Training accuracies were taken from the final epoch. 
The best test accuracy was achieved by the combination of Swish and Mish --- it was equal to 0.8718.
Second best performance was achieved by \functionName{} + Mish.
Combination originally used by Afchar et al., ReLU + LeakyRelu, was characterized by a slightly worse performance --- 0.8676.

As mentioned earlier, Afchar et al. provided pretrained weights of their architectures --- these models were also evaluated in the course of the experiment. 
They achieved the accuracies of 0.9513 and 0.9131 respectively for training and testing sets.
These values differ from the ones the ones presented in Table~\ref{tab:meso-net-third-stage-results}. 
The reason behind that are mainly the differences between training processes. 
As some of the details like number of epochs or types and degrees of augmentations were not provided, we could not exactly reproduce their training procedure.

Our evaluation process covered not only the performance of activation function but also times of inference.
Treating ReLU + ReLU as a baseline, we obtained +2.2\% for Swish, +2.3\% for Leaky ReLU, +4.3\% for Mish, and +7.5\% for \functionName{}. Note that, since~\cite{mesonet} did not provide the exact time requirements of their training and inference, hence we need to treat those as a reference.

Described experiment shows that use of newer activation functions can indeed improve the performance of the neural network. 
New activation function introduced in this paper --- \functionName{}, achieved results that were comparable to state--of--the--art functions.

Despite being slightly more demanding in the terms of required computations, the provided results showed that it is a promising and reliably consistent alternative to the existing solutions and that it is worth to explore further its characteristics.

\section{Additional \functionName{} tests}\label{sect:additional-tests}

In~\cite{mish}, aside from proposing Mish activation function, extensive tests and benchmarks were provided in the related repository~\cite{mish_repository}. 
The following section contains the result gathered from four of these benchmarks that were later used to compare \functionName{} with well--known activation functions.

\subsection{Various depths of the neural network}

First test aimed to give information about the performance of neural network in relation to its depth.
The architectures used in the benchmark differed only in the number of fully connected layers. 
The only difference between the benchmark used in~\cite{mish} and the one presented in the paper was the dataset used --- instead on MNIST~\cite{mnist}, activation functions were evaluated on Fashion MNIST dataset~\cite{fashion_mnist}.
Both sets consist of 60,000 training and 10,000 test gray--scale images of size $28\times28$. However, samples from Fashion MNIST contain more degrees of freedom making it more difficult. Moreover, due to various textures and intensity changes as well as actually appearing in Deepfake videos, the clothing from Fashion MNIST seem more relevant than the digits from MNIST.

Each network started with two convolutional layers, followed by a max pooling with a dropout of rate 25\%. It was followed by a number of fully connected layers of 500 neurons with batch normalization and dropout (once again 25\%). The quantity of layers depended on the parameter which was from range from 15 to 25. 
Final layer of the network was composed of output neurons using softmax activation function.
Both convolutional and fully--connected layers used the currently evaluated activation function. Discussed benchmark was performed using ReLU, Swish, Mish and \functionName{} functions. 
The optimizer utilized was stochastic gradient descent (SGD), all runs used the same learning rate with a batch size of 128.
We ran the process for 50 epochs, similarly to the experiment described in Sect.~\ref{sect:methodology}.

\begin{figure}[h]
    \begin{center}
      \includegraphics[width=0.9\linewidth]{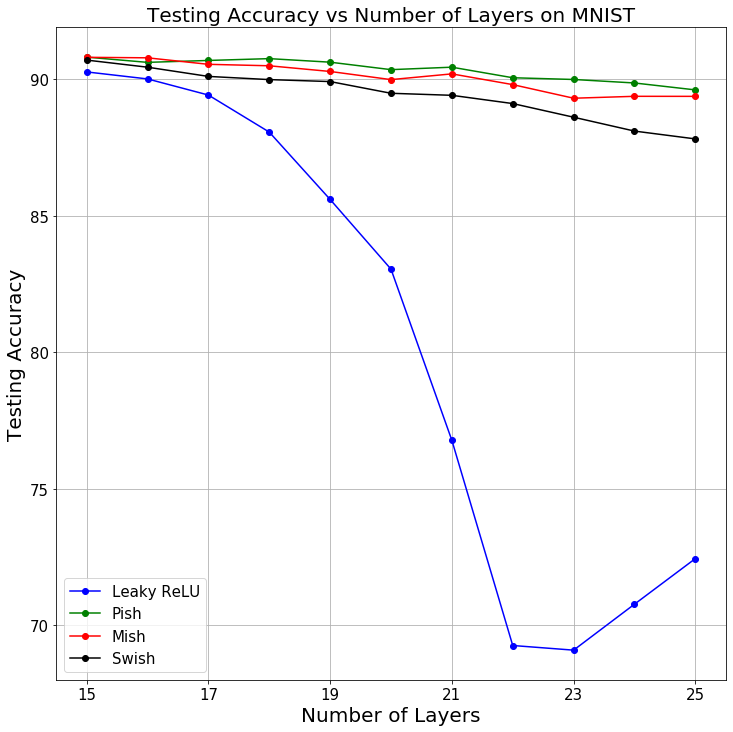}
    \end{center}
      \caption{Maximum test accuracy in relation to the depth of the neural network. Plot generated using a script from~\cite{mish_repository}.}
    \label{fig:neural_network_depth_comparison} 
\end{figure}

Figure~\ref{fig:neural_network_depth_comparison} shows test accuracies of aforementioned activation functions. Presented results are the maximum test accuracies scored for each consecutive number of layers.
Note that, the values remained similar for the smaller number of layers. As their number increased, so had the differences between functions' performance. 
While the differences are not big, \functionName{} achieves the best test accuracy in the case of almost all network's depths.

\subsection{SqueezeNet}\label{ssect:squeezenet}

The second benchmark concerned evaluation of popular architecture SqueezeNet~\cite{squeeze-net}.

Described process was performed using CIFAR--10~\cite{cifar-10}, a dataset composed of 60,000 samples of size of $32\times32$ divided equally into 10 classes.
All activation functions (ReLU, Mish, Swish and \functionName{}) were trained for 100 epochs using the same parameters --- batch size of 128 and a learning rate equal of $10^{-3}$.
Process was repeated 3 times to ensure the reliability of the results.

\begin{table}[h]
    \begin{center}
    \begin{tabular}{|l|c|}
    \hline
    Activation function & Top--1 accuracy \\
    \hline\hline
    Swish               & \textbf{0.8874} \\ 
    \functionName{}     & 0.8860          \\ 
    Mish                & 0.8845          \\ 
    ReLU                & 0.8784          \\ 
    \hline
    \end{tabular}
\end{center}
    \caption{Top--1 test accuracies of SqueezeNet neural network trained on CIFAR--10 dataset}
    \label{tab:squeezenet-top1acc}
\end{table}

Table~\ref{tab:squeezenet-top1acc} contains top--1 test accuracies of SqueezeNet trained using ReLU, Mish, Swish and \functionName{} functions. Models were evaluated after each epoch --- presented values are the maximum achieved by each of the networks.

The results of the well known functions are as expected. ReLU function provided smallest accuracy, about 0.01 less than the other functions, which is a typical outcome confirmed by many benchmarks. Meanwhile, results of Mish and Swish are close, respectively 0.8874 and 0.8844 --- as stated in the original Mish paper, there are cases when Mish function performs slightly worse than the other one.
The second best score, 0.8860, belonged to \functionName{}.

\subsection{Other architectures}

\functionName{} has been also tested on other architectures like EfficientNet~\cite{efficientnet} and DenseNet~\cite{dense-net}. 
Their evaluation procedures were similar to the one described in Sect.~\ref{ssect:squeezenet} --- models were trained on CIFAR--10 dataset.
ReLU, Mish, Swish and \functionName{} scored maximally respectively 0.8026, 0.8043, 0.8035, 0.7875 for EfficientNet and 0.9103, 0.9112, 0.9132, 0.9089 for DenseNet.

\section{Local Feature Descriptors}\label{sect:lfd}

Akhtar et al. in their paper \cite{local_feature_descriptors} showed their preliminary results of detecting Deepfake manipulation with the use of local feature descriptors.

The following section contains results of our preliminary comparison between two of the methods from~\cite{local_feature_descriptors} and two descriptors which were not previously investigated in the field of Deepfake detection.

Presented experiment was performed on the subset of FaceForensics++ dataset \cite{faceforensics++}.
It was composed of 150 original and 150 corresponding Deepfake videos. 
80\% of movies were used in training set, whereas the rest was used in test. 
Training was run on each frame extracted from every video.

Classification between pristine and manipulated samples was done with images' keypoints. 
They were obtained using SIFT \cite{sift}, SURF \cite{surf}, ORB \cite{orb} and BRISK \cite{brisk} feature descriptors --- latter two were not covered in the original work. 
The procedure used, was inspired by the approach utilized by Akhtar et al., however the following experiment was created using our own implementation. This led to some differences between these two approaches. 
In our experiments each feature descriptor used only its coordinates and angles creating feature vectors of 128 values, moreover we decided to use more common dataset --- FaceForensics++.

The pipeline worked as follows --- it started with localization of the face using Dlib face detector \cite{dlib}.
The basis of its bounding boxes image was cropped and later normalized to the size of $200\times200$. 
Next step concerned keypoints' extraction. We used only information about their coordinates and angle. 
In order to further normalize the data, whole frame was split into 16 equally sized chunks ($4\times4$). Each keypoint was assigned to the appropriate chunk basing on its coordinates.
Each of the chunks was later converted into a histogram of 8 bins. The assignment was done using the angles of the points (first 45 degrees were moved to the first bin, and so on). This resulted in a vector of 128 values --- 8 bins for each of the 16  chunks.
Such data was later used in classification. In order to make solution simple, we utilized support vector machine with no hyperparameter optimization.

\begin{table}[h]
\begin{center}
\begin{tabular}{|l|c|c|c|}
\hline
Descriptor & Accuracy & Recall & EER \\
\hline\hline
SIFT~\cite{local_feature_descriptors} [LQ] & - & - &0.5710 \\
SURF~\cite{local_feature_descriptors} [LQ] & - & - & 0.4546 \\
SIFT~\cite{local_feature_descriptors} [HQ] & - & - &0.5758 \\
SURF~\cite{local_feature_descriptors} [HQ] & - & - & 0.6726 \\
SURF       & 0.6105 & 0.5996 & 0.3917 \\
SIFT       & 0.6742 & 0.6314 & 0.3395 \\
ORB        & 0.7059 & 0.6754 & 0.3059 \\
BRISK      & 0.7279 & 0.7012 & 0.2836 \\
\hline
\end{tabular}
\end{center}
\caption{\label{tab:lfd_results}Results achieved by the feature descriptors method on a subset of FaceForensics++ dataset. Note that referenced results did not provide Accuracy and Recall, the authors used a different dataset as well.}
\end{table}

Table~\ref{tab:lfd_results} contains test results achieved by the aforementioned feature descriptors. Just like in the case of Akhtar et al. we treated each frame as a distinct sample --- other frames of the video were not taken into account when testing a single sample. In a spirit of detecting Deepfakes, we treated manipulated samples as positives and pristine as negatives.
The use of BRISK and ORB methods contributed to the improvement of all metrics.
The change of feature descriptor has increased the accuracies from about 9 to 12\% in case of SURF and from 3 to 5\% in case of BRISK.

\paragraph{Time comparison with MesoNet.} Since our main goal is to provide an efficient, yet fast method of Deepfake detection, performable on office--grade PCs, our evaluation process has also included the time evaluation of both training and inference of MesoNet and local feature descriptors approaches. 
As two solutions were trained on different datasets, we compared their training times by running this procedure once again on MesoNet training set which consisted of about 12.500 images. MesoNet's model was trained for 10 epochs, as this quantity contributed to the satisfying results. 
Time measurement of LFD--based approach took into consideration both feature extraction (creation of feature vectors) and training of the SVM classifier.

Training process of MesoNet network on this dataset took about 30 minutes on consumer grade GPU, whereas LFD approach required about 3.5 minutes without GPU, utilizing only the CPU (60\% of this time was required to convert frames into feature vectors using feature descriptors).
Inference based on an average of 5000 predictions was equal to respectively 0.034 seconds for MesoNet and 0.076 for LFD.
The reason behind greater time of descriptor--based approach was preprocessing which converted images to feature vectors taking about 98\% of inference time.

\section{Conclusion} \label{sect:conclusions}

In the paper we focus on Deepfake detection, that becomes a serious threat to the privacy and biometric security. We put particular interest in providing a methods of detecting Deepfakes that are feasible for consumer grade devices, so that anyone can verify, with reasonable probability, whether the video is legitimate. Our contribution is twofold -- we investigated the efficacy of a shallow neural network MesoNet, with various activation functions. Aside experimental verification  of previously introduced function, we presented a novel one -- \functionName{}{}, that achieves results competitive to top-performing ones, yet with high consistency. Using MesoNet with \functionName{} and Mish activation functions allowed us to achieve $0.8689$ test accuracy, with only Swish and Mish combination outperforming it (over $0.87$) and the baseline combination (ReLU+Leaky ReLU) trailing  with $0.8676$.
The other part of the contribution is an extension of Deepfake detection method based on local feature descriptors. Due to using location and angle of BRISK features we obtained Equal Error Rate of $0.28$. In the future work, we plan to focus on expanding the LFD--based methods, to achieve lower EER, as well as further lowering time and computation requirements of the Deepfake detection. 

{\small
\bibliographystyle{plain}
\bibliography{bibliography}

\begin{thebibliography}{10}

\bibitem{mish_repository}
{digantamisra98} mish.
\newblock \url{https://github.com/digantamisra98/Mish}.
\newblock Accessed: 15-05-2020.

\bibitem{psd2}
{European Parliament} directive (eu) 2015/2366.
\newblock
  \url{https://eur-lex.europa.eu/legal-content/EN/TXT/PDF/?uri=CELEX:32015L2366&from=EN}.
\newblock Accessed: 13-03-2020.

\bibitem{mesonet}
Darius Afchar, Vincent Nozick, Junichi Yamagishi, and Isao Echizen.
\newblock Mesonet: a compact facial video forgery detection network.
\newblock {\em CoRR}, abs/1809.00888, 2018.

\bibitem{local_feature_descriptors}
Z.~{Akhtar} and D.~{Dasgupta}.
\newblock A comparative evaluation of local feature descriptors for deepfakes
  detection.
\newblock In {\em 2019 IEEE International Symposium on Technologies for
  Homeland Security (HST)}, pages 1--5, 2019.

\bibitem{surf}
Herbert Bay, Tinne Tuytelaars, and Luc Van~Gool.
\newblock Surf: Speeded up robust features.
\newblock In Ale{\v{s}} Leonardis, Horst Bischof, and Axel Pinz, editors, {\em
  Computer Vision -- ECCV 2006}, pages 404--417, Berlin, Heidelberg, 2006.
  Springer Berlin Heidelberg.

\bibitem{xception-net}
Fran{\c{c}}ois Chollet.
\newblock Xception: Deep learning with depthwise separable convolutions.
\newblock {\em CoRR}, abs/1610.02357, 2016.

\bibitem{frequency_analysis}
Ricard Durall, Margret Keuper, Franz-Josef Pfreundt, and Janis Keuper.
\newblock Unmasking deepfakes with simple features, 2019.

\bibitem{relu}
Xavier Glorot, Antoine Bordes, and Y.~Bengio.
\newblock Deep sparse rectifier neural networks.
\newblock {\em Proceedings of the 14th International Conference on Artificial
  Intelligence and Statisitics (AISTATS) 2011}, 15:315--323, 01 2011.

\bibitem{gan}
Ian~J. Goodfellow, Jean Pouget-Abadie, Mehdi Mirza, Bing Xu, David
  Warde-Farley, Sherjil Ozair, Aaron Courville, and Yoshua Bengio.
\newblock Generative adversarial networks, 2014.

\bibitem{resnet-50}
Kaiming He, Xiangyu Zhang, Shaoqing Ren, and Jian Sun.
\newblock Deep residual learning for image recognition.
\newblock {\em CoRR}, abs/1512.03385, 2015.

\bibitem{dense-net}
G.~{Huang}, Z.~{Liu}, L.~{Van Der Maaten}, and K.~Q. {Weinberger}.
\newblock Densely connected convolutional networks.
\newblock In {\em 2017 IEEE Conference on Computer Vision and Pattern
  Recognition (CVPR)}, pages 2261--2269, 2017.

\bibitem{squeeze-net}
Forrest~N. Iandola, Matthew~W. Moskewicz, Khalid Ashraf, Song Han, William~J.
  Dally, and Kurt Keutzer.
\newblock Squeezenet: Alexnet-level accuracy with 50x fewer parameters and
  {\textless}1mb model size.
\newblock {\em CoRR}, abs/1602.07360, 2016.

\bibitem{dlib}
Davis~E. King.
\newblock Dlib-ml: A machine learning toolkit.
\newblock {\em Journal of Machine Learning Research}, 10:1755--1758, 2009.

\bibitem{adam}
Diederik~P. Kingma and Jimmy Ba.
\newblock Adam: A method for stochastic optimization.
\newblock {\em CoRR}, abs/1412.6980, 2015.

\bibitem{korshunov}
Pavel Korshunov and Sebastien Marcel.
\newblock Deepfakes: a new threat to face recognition? assessment and
  detection, 2018.

\bibitem{cifar-10}
Alex Krizhevsky.
\newblock Learning multiple layers of features from tiny images.
\newblock {\em University of Toronto}, 05 2012.

\bibitem{mnist}
Y.~{Lecun}, L.~{Bottou}, Y.~{Bengio}, and P.~{Haffner}.
\newblock Gradient-based learning applied to document recognition.
\newblock {\em Proceedings of the IEEE}, 86(11):2278--2324, 1998.

\bibitem{brisk}
Stefan Leutenegger, Margarita Chli, and Roland Siegwart.
\newblock Brisk: Binary robust invariant scalable keypoints.
\newblock pages 2548--2555, 11 2011.

\bibitem{sift}
David Lowe.
\newblock Distinctive image features from scale-invariant keypoints.
\newblock {\em International Journal of Computer Vision}, 60:91--, 11 2004.

\bibitem{deepfake_survey_2}
Siwei Lyu.
\newblock Deepfake detection: Current challenges and next steps, 2020.

\bibitem{leaky_relu}
Andrew~L. Maas, Awni~Y. Hannun, and Andrew~Y. Ng.
\newblock Rectifier nonlinearities improve neural network acoustic models.
\newblock In {\em in ICML Workshop on Deep Learning for Audio, Speech and
  Language Processing}, 2013.

\bibitem{mish}
Diganta Misra.
\newblock Mish: A self regularized non-monotonic neural activation function,
  2019.

\bibitem{deepfake_survey_1}
Thanh~Thi Nguyen, Cuong~M. Nguyen, Dung~Tien Nguyen, Duc~Thanh Nguyen, and
  Saeid Nahavandi.
\newblock Deep learning for deepfakes creation and detection, 2019.

\bibitem{swish}
Prajit Ramachandran, Barret Zoph, and Quoc~V. Le.
\newblock Searching for activation functions, 2017.

\bibitem{faceforensics++}
Andreas R\"ossler, Davide Cozzolino, Luisa Verdoliva, Christian Riess, Justus
  Thies, and Matthias Nie{\ss}ner.
\newblock Face{F}orensics++: Learning to detect manipulated facial images.
\newblock In {\em International Conference on Computer Vision (ICCV)}, 2019.

\bibitem{orb}
E.~{Rublee}, V.~{Rabaud}, K.~{Konolige}, and G.~{Bradski}.
\newblock Orb: An efficient alternative to sift or surf.
\newblock In {\em 2011 International Conference on Computer Vision}, pages
  2564--2571, 2011.

\bibitem{inception_layer}
Christian Szegedy, Wei Liu, Yangqing Jia, Pierre Sermanet, Scott~E. Reed,
  Dragomir Anguelov, Dumitru Erhan, Vincent Vanhoucke, and Andrew Rabinovich.
\newblock Going deeper with convolutions.
\newblock {\em CoRR}, abs/1409.4842, 2014.

\bibitem{efficientnet}
Mingxing Tan and Quoc~V. Le.
\newblock Efficientnet: Rethinking model scaling for convolutional neural
  networks, 2019.

\bibitem{fashion_mnist}
Han Xiao, Kashif Rasul, and Roland Vollgraf.
\newblock Fashion-mnist: a novel image dataset for benchmarking machine
  learning algorithms, 2017.

\bibitem{head_pose_estimation}
Xin Yang, Yuezun Li, and Siwei Lyu.
\newblock Exposing deep fakes using inconsistent head poses.
\newblock {\em CoRR}, abs/1811.00661, 2018.

\end{thebibliography}
}

\end{document}